%% file: main.tex
\documentclass{article}

\usepackage{microtype}
\usepackage{graphicx}
\usepackage{subfigure}
\usepackage{booktabs} 
\usepackage{lipsum}
\usepackage{algorithm}
\usepackage{algorithmic}
\usepackage[inline]{enumitem}

\usepackage{defs}
\usepackage{hyperref}

\setlist[itemize]{noitemsep,topsep=-6pt}
\setlist[enumerate]{noitemsep,topsep=-6pt}

\usepackage[accepted]{icml2021}

\icmltitlerunning{Generative Video Transformer}

\usepackage{amsmath}
\usepackage{amsfonts}

\begin{document}

\twocolumn[
\icmltitle{Generative Video Transformer: Can Objects be the Words?}



\icmlsetsymbol{equal}{*}

\begin{icmlauthorlist}
\icmlauthor{Yi-Fu Wu}{rutgers}
\icmlauthor{Jaesik Yoon}{rutgers,sap}
\icmlauthor{Sungjin Ahn}{rutgers,ruccs}
\end{icmlauthorlist}

\icmlaffiliation{rutgers}{Department of Computer Science, Rutgers University}
\icmlaffiliation{sap}{SAP Labs}
\icmlaffiliation{ruccs}{Rutgers Center for Cognitive Science}

\icmlcorrespondingauthor{Yi-Fu Wu}{yifu.wu@gmail.com}
\icmlcorrespondingauthor{Sungjin Ahn}{sjn.ahn@gmail.com}

\icmlkeywords{Machine Learning, ICML}

\vskip 0.3in
]



\printAffiliationsAndNotice{} 

\begin{abstract}

Transformers have been successful for many natural language processing tasks.
However, applying transformers to the video domain for tasks such as long-term video generation and scene understanding has remained elusive due to the high computational complexity and the lack of natural tokenization.
In this paper, we propose the Object-Centric Video Transformer (OCVT) which utilizes an object-centric approach for decomposing scenes into tokens suitable for use in a generative video transformer.
By factoring the video into objects, our fully unsupervised model is able to learn complex spatio-temporal dynamics of multiple interacting objects in a scene and generate future frames of the video.
Our model is also significantly more memory-efficient than pixel-based models and thus able to train on videos of length up to 70 frames with a single 48GB GPU.
We compare our model with previous RNN-based approaches as well as other possible video transformer baselines. We demonstrate OCVT performs well when compared to baselines in generating future frames. OCVT also develops useful representations for video reasoning, achieving start-of-the-art performance on the CATER task.

\end{abstract}

\section{Introduction}
Recent advances in natural language processing (NLP) have shown that models trained with an autoregressive language modeling objective using large transformers \citep{transformer} can learn to generate realistic text passages \citep{gpt, gpt2, gpt3}.
Furthermore, the representations learned with these generative pre-trained (GPT) models are effective at downstream tasks such as question answering, machine translation, reading comprehension, and summarization. While it is of primary interest to develop an analogous generative pre-training procedure for videos, the computational overhead in dealing with videos has made this a difficult endeavor.





The main challenges in developing video transformers are (1) how to tokenize a video and (2) how to serialize the tokens because unlike text the ordered symbolic structure is not naturally given in videos.
Several previous attempts for images \citep{igpt, imagetransformer} and videos \citep{videotransformer}
operate at the pixel level, flattening out an image into a sequence of pixels. However, since the memory and computation of the self-attention layers used in transformers are quadratic in the input sequence length, in order to train these models efficiently, these works either lower the resolution of the image or use local attention instead of global attention across the entire image or video.

In this paper, we investigate a potentially different approach to tackling this quadratic cost. We leverage the inductive bias that our world is made of objects and working at an object-level granularity can be beneficial 
for many downstream tasks, especially in scenes where the objects interact with each other. 
To this end, we investigate different design choices for tokenizing and serializing a video and propose the Object-Centric Video Transformer (\oursns). In OCVT, we combine object-centric representations with a transformer trained using an autoregressive object-level next-frame prediction objective.
Our model leverages a class of object-centric latent representations~\citep{air,spair,space,gswm} which can learn structured representations without object-level labeling.
The learned object representation includes explicit location and size information about the objects in each video frame. We use this to find a bipartite matching of objects between frames, allowing us to construct the object-wise loss function.
The use of an object-centric transformer allows \ours to learn spatial and long-term temporal interactions between objects in a video.

We evaluate \ours in a number of environments constructed to demonstrate the strengths and limitations of the model.
Given a number of initial ground-truth video frames, \ours is able to generate future predictions of a video, even in scenarios where the dynamics of the objects in the video depend on interactions made many frames in the past.
Lastly, the representations \ours learns are also able to handle downstream video understanding tasks that require long-term spatial-temporal reasoning, achieving results comparable with the state of the art on the CATER \citep{cater} snitch localization task.

\section{Related Works}

\textbf{Unsupervised Object-Centric Representation.}
Recent advances in unsupervised object-centric representation learning for images can be split into two approaches.
Scene-mixture models \citep{nem, rnem, monet, iodine, genesis, slotattention} decompose a scene into objects using a mixture of image-sized components that are each generated by a distributed representation.
Bounding box methods \citep{air, spair} use spatial attention to explicitly obtain object position and size information.
SPACE \citep{space} combines these two approaches by obtaining both explicit bounding boxes for objects as well as image-sized object segmentation masks for parts of the image that cannot be cleanly captured by bounding boxes.
Previous works that extend these models to videos utilize an RNN for temporal modeling \citep{op3, cobra, sqair, silot, scalor, gswm}. Several of these models, namely STOVE \citep{stove} and GSWM \citep{gswm}, also model interactions among objects with a graph neural network.

\textbf{Transformers for Images \& Videos.}
Several recent works have applied transformer-based architectures to various tasks for visual scenes.
To handle the quadratic memory and computation cost in the number of pixels of an image, \citet{imagetransformer} restrict the self-attention mechanism to attend to local neighborhoods instead of the entire image. 
\citet{videotransformer} extend this technique to videos.
Other attempts at solving this quadratic cost lower reduce the resolution of the image \citep{igpt}, use approximations of global attention \citep{sparse_transformers}, restrict self-attention along an axis \citep{axial1, axial2}, or work with patches of the original image \citep{vit}.
Other works operate in the latent space instead of directly on the image pixels.
DETR \citep{detr} uses the convolutional feature map as input to the transformer for effective object detection and Trackformer \citep{trackformer} extends this technique to videos for object tracking. \citet{latentvt} use a VQ-VAE \citep{vqvae} to obtain discrete latent representations before applying the transformer in the latent space.

\textbf{Object-centric Approaches with Transformers.} 
There have also been several recent attempts at combining object-centric representations with transformers.
Hopper \citep{hopper} uses object-centric representations obtained by DETR \citep{detr} in a multi-hop transformer for spatio-temporal reasoning and is applied to the CATER snitch localization task.
In addition to requiring supervised object labels (bounding boxes) for the objects, Hopper also uses several auxiliary losses that require knowing the first and second movements of the snitch.
AlignNet \citep{alignnet2, alignnet} uses MONet~\citep{monet} for object-centric representations and leverages a transformer's attention matrix to align objects between frames to track objects over a video.
While AlignNet provides good performance for tracking tasks that require modeling of object permanence, it is not experimented on any tasks that require the generation of future frames.
Concurrent with our work, Objects-Align-Transition \citep{oat} extends AlignNet by including a transition model to perform future generation.
\citet{monetbert} also use MONet to obtain object-centric representations and input the learned representations into a transformer.
In addition to the supervised loss for the particular task, a self-supervised BERT-style \citep{bert} masked prediction loss is included.
While the self-supervised loss helps to improve the model's performance in downstream tasks, without autoregressive modeling, this model cannot generate future frames.

\section{Object-Centric Video Transformer}


%
%
%
%
%
%
%
%
%
%









\subsection{Key Ideas}

\subsubsection{Discretizing Video}
To design a generative video transformer, we must first decide how to appropriately tokenize a video into discrete entities to use as input to a transformer.
While the discrete structure of words or sub-words provides a natural choice for language modeling, it is not obvious what analogous tokenization is for video.

One option is to consider each image as a token.
However, without a properly disentangled representation, this choice may limit the ability of the transformer to model interactions within an image.
This would be analogous to sentence-level tokenization in text, where the encoding of each sentence, e.g. using an RNN, is used as input to the transformer.
The other extreme is to consider each pixel as a token.
However, the quadratic memory and computation cost of transformers in sequence length would limit the ability of such a model to work on long videos.

Therefore, we arrive at the following observations about the proper tokenization to use for a generative video transformer: the area for each visual discrete entity should (1) cover as large of a pixel area as possible for computational efficiency, but (2) without being too large as to prevent the modeling of interactions between different parts of the image.

One possible approach to such a middle ground is to use convolutional feature maps as a tokenization strategy for each image of the video. 
In this work, we investigate an object-centric approach, tokenizing an image into its constituent object representations.
We argue that this is a reasonable choice for videos because in the physical world, the spatiotemporal dynamics of a scene is governed mostly by the causal interaction among the objects of the scene.
It is also true that an object may change its state rather independently of other objects, e.g., clock hands or a walking person, but the updated state would still be dependent on its previous states.
Thus, we hypothesize that accessing such historical states at the individual object-level would also be beneficial compared to doing it at pixel, image, or feature map levels where the identity of the objects is not necessarily preserved in the representation across frames.


\begin{figure*}[t]
    \hspace{-6mm}
    \centering
    \includegraphics[width=0.85\linewidth]{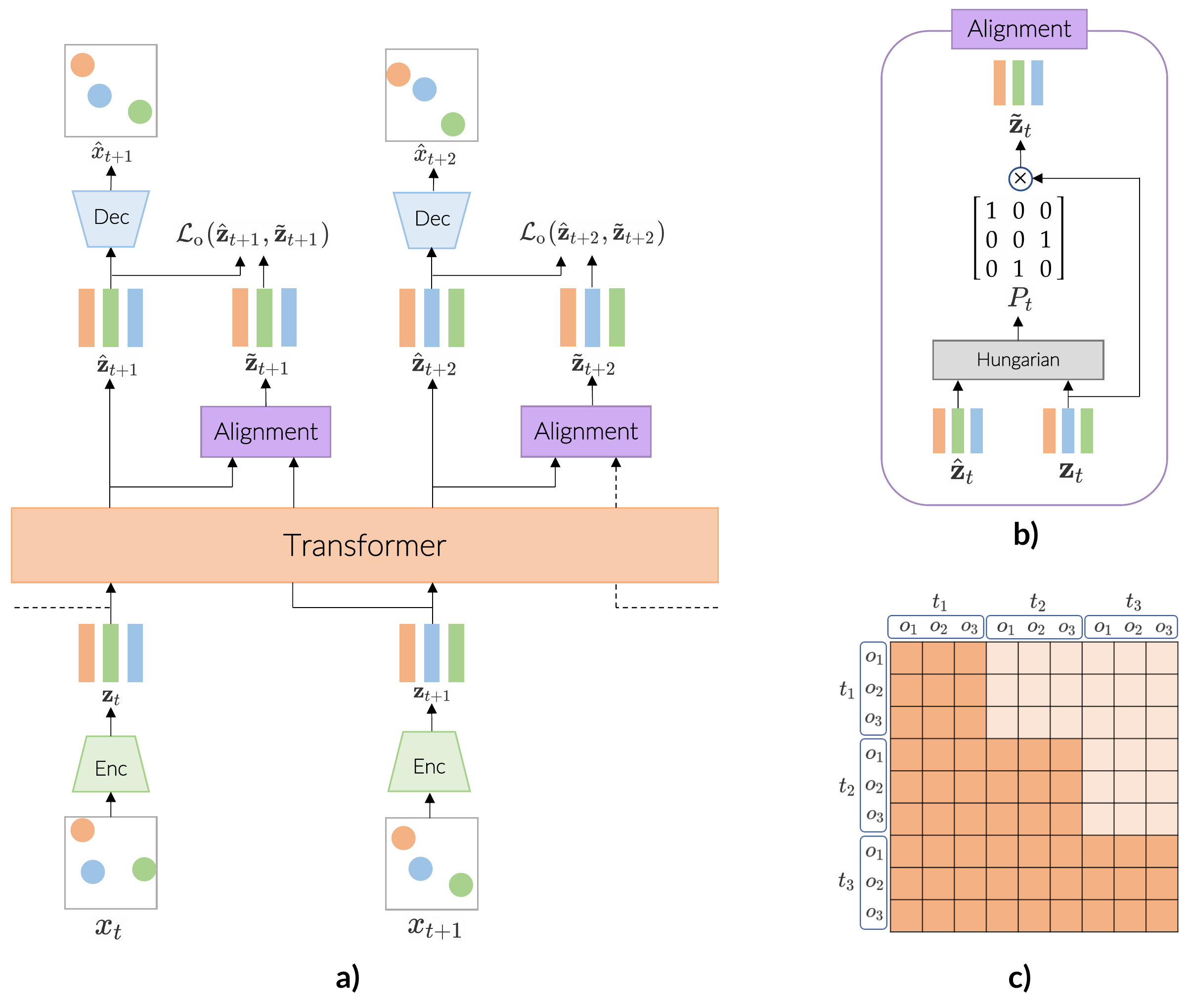}
    \vspace{-5mm}
    \caption{(a) Overview of model architecture. Object-centric representations of a video are used in a transformer to predict future frames in the latent space. A alignment algorithm is used to ensure that an object-wise loss can be used between frames. (b) The alignment algorithm uses the Hungarian algorithm obtain a permutation matrix, which is then multiplied by $\bz_t$ to obtained the aligned latents $\tilde{\bz}_t$. (c) The transformer's causal self-attention mask allows objects within the same timestep to attend to each other.}
    \label{fig:model}
    \vspace{-3mm}
\end{figure*}

\subsubsection{Serializing Objects}
Continuing the analogy with language modeling, if we consider objects as visual words, we may consider an image to be a visual sentence.
However, unlike words in a sentence, objects in an image do not have a natural order because they consist of a \textit{set} of object representations, not a sequence.
Therefore, unlike the serialization of words in text, we cannot simply concatenate the object representations for use as input to the transformer.

Suppose we were able to impose an artificial order for the objects in the image, e.g. raster scan order based on object positions.
Then we would be able to concatenate the sorted objects in a predefined way and predict each object autoregressively.
Since the generation of each object would depend on previous objects in the scene, this results in a flexible model capable of modeling dependencies within objects in the scene.
However, in dynamic scenes where the object positions change over time with complex interactions, predicting the correct object ordering may not be an easy task to learn~\cite{ordermatters, sqair}.
Further, such a model would scale linearly with the number of objects in an image and thus may not be practical for scenes with many objects.

A different approach that we leverage in this work is to predict the entire image at once by generating all the objects in the image simultaneously given their previous states.
The autoregressive prediction objective in this case would be an object-wise loss between subsequent video frames.
Instead of requiring the model to learn an arbitrary ordering of the set of objects, we would only need to correctly align the objects in adjacent frames so that the object-wise loss can be applied.
This alignment can be done based on object location by leveraging object representations with explicit position and size information of the objects, such as those given by SPACE \citep{space}.
By generating an entire image in one forward pass through the transformer, generation time for each video frame is constant with respect to the number of objects in a scene.
Furthermore, this choice also provides a natural objective for future prediction in physical scenes as the object-wise loss forces the model to learn the dynamics for how objects change over time.

\subsection{Architecture}

We now describe the overall architecture for our proposed model, the Object-Centric Video Transformer (OCVT). 
Given a length-$T$ video $\bx_{1:T}=(\bx_1, \ldots, \bx_T)$, we use an encoder to produce a set of latents $\bz_{1:T}$.
This encoder is part of a structured VAE designed so that $\bz_{1:T}$ are object-centric representations of the scene with explicit bounding box information.
These latents are used as input to a transformer decoder to obtain $\hat{\bz}_{2:T+1}$, the predicted latents at the next timestep.
In order to align the objects in $\hat{\bz}_t$ with the objects in $\bz_t$, we obtain a permutation matrix $P_t$ by leveraging the Hungarian algorithm with a cost matrix that consists of the pairwise matching cost between each object in $\bz_t$ and $\hat{\bz}_t$.
This permutation matrix is then multiplied with $\bz_t$ to produce $\tilde{\bz}_t$, which is aligned with $\hat{\bz}_t$.
After alignment, we can then train our model with an object-wise loss between $\hat{\bz}_t$ and $\tilde{\bz}_t$.
To generate a reconstruction of the image, we input $\hat{\bz}_t$ into the decoder of the structured VAE to obtain $\hat{\bx}_t$. 
This architecture is depicted in Figure \ref{fig:model}a.
We now describe each of these components in more detail.

\subsubsection{Object-Centric Representations}
As described earlier, in order to correctly align objects between frames, it is crucial for the object-centric representations in our model to contain explicit location and size information.
Thus, we leverage a structured VAE similar to SPACE \citep{space}.
The encoder of the VAE consists of a fully convolutional network that transforms the input image to a grid of $H \times W$ cells.
The feature map of each grid cell is then run through another fully convolutional network to produce the latents $\bz$.
Each latent variable $\bz$ consists of four components $\bz=[\bz^\pres, \bz^\where, \bz^\depth, \bz^\what]$.
$\bz^\pres \in \{0,1\}$ is a binary random variable denoting the presence of an object, $\bz^\where=[z^h, z^w, z^x, z^y]$ represents the bounding box and center location of the object, $\bz^\depth \in \mathbb{R}$ specifies the depth of an object, and $\bz^\what$ is a representation for everything else about the object (e.g., appearance).

We do this for each timestep $t$ resulting in one latent representation $\bz_t$ for each image in the video $\bx_t$.
Note that since each grid cell produces a set of latents for an object, we are able to detect up to $H \times W$ objects in one image.
In our experiments, we choose $H \times W$ to be larger than the total number of objects $O$ and use $\bz^\pres$ to determine whether or not a grid cell detects an object.

The decoder uses a series of deconvolutional layers to create an image for each object.
$\bz^\pres$ is used to determine the transparency of the object.
A low $\bz^\pres$ would result in an object not appearing in the reconstructed image.
A spatial transformer \citep{spatial_transformer} is then used with $\bz^\where$ to place each object onto the final reconstructed image.
For scenes with a background that cannot be completely captured by objects, we also train a fully convolutional background module to generate a background latent $\bz^\bg$. In this case, we also generate a foreground mask $\alpha$ that controls the weighting between the foreground objects and the background in the final rendered image.



We pretrain the encoder and decoder networks on the video frames and freeze the weights of the networks when the transformer is being trained.
Full implementation and training details can be found in the Supplementary Material.

\subsubsection{Object-Centric Transformer}
In order to model the dynamics of the objects over time, we use a transformer decoder where the inputs are the object-centric latents $\bz_t$.
Compared to using an RNN as is done in other models \citep{sqair, silot, scalor, gswm}, the transformer can model both dynamics of an object over time as well as interactions between objects without requiring a separate interaction module.
Moreover, while RNNs store information from past states in their hidden state, transformers have direct access to states in the past, allowing for better modeling of long-term dependencies.

In addition to $\bz_t$, we also use a sinusoidal encoding for the timestep $t$, similar to the positional encoding in the original transformer \citep{transformer} as input to the transformer decoder.
Furthermore, we modify the causal attention mask from the traditional transformer decoder so that the objects within a timestep can attend to each other (Figure \ref{fig:model}c).

After the final transformer block, we run the output through a single hidden layer MLP to produce $\hat{\bz}_{t+1}$, the predicted object representations for the next timestep.
$\hat{\bz}_{t+1}^\what$, $\hat{\bz}_{t+1}^\depth$, and $\hat{\bz}_{t+1}^\pres$ are predicted directly as the output of the MLP, but for the bounding box of the object $\hat{\bz}_{t+1}^\where$, we predict an update $\Delta \hat{\bz}_{t+1}^\where$.
We then apply the following formula to obtain the predicted bounding box: $\hat{\bz}_{t+1}^\where = \bz_t^\where + c \cdot \text{tanh}(\Delta \hat{\bz}_{t+1}^\where)$, where $c$ is a hyperparameter between 0 and 1 controlling the maximum update in one timestep.
This choice reflects the fact that objects generally do not change size and location significantly from one timestep to the next.
Additionally, since the cost matrix used for alignment (discussed in the next section) is based on the distance between the predicted bounding box and the actual bounding box, centering $\hat{\bz}_{t+1}^\where$ around the previous bounding box $\bz_{t}^\where$ essentially initializes the alignment to be between objects that are the closest between two timesteps, which in most cases would be the correct alignment.
This facilities training early on before the model has learned the correct dynamics of the objects.






\subsubsection{Object Alignment} As an object moves around in the image, it may end up being detected by different grid cells at different time steps.
This means that the objects in one timestep may not be aligned with the same objects at other timesteps.
As discussed earlier, alignment is necessary for our model because we use an object-wise loss between frames.
That is, we need the objects in our VAE inferred latents $\bz_t$ to be in the same order as the latents predicted by the transformer $\hat{\bz_t}$.
It should be noted that the input of the transformer \textit{does not} need to be aligned across timesteps because the object-wise self-attention works on an order-less set of objects.

In order to align the objects between timesteps, we leverage the Hungarian algorithm \citep{hungarian}, which solves the bipartite matching problem given a cost matrix in polynomial time. 
The explicit position and size information from the latent $\bz^\where$ makes it easy to define a position-based cost function for the matching of objects in adjacent frames.
Let us denote $\bz_{t,k}$ as the latent corresponding to grid cell $k$ at timestep $t$.
We construct a cost matrix that consists of the pairwise matching cost $\Lmatch(\hat{\bz}_{t,i}, \bz_{t,j})$ between the transformer predicted latents $\hat{\bz}_{t,i}$ and the VAE inferred latents $\bz_{t,j}$.
$\Lmatch$ is defined as: $\Lmatch(\hat{\bz}_{t,i}, \bz_{t,j}) =$
\begin{align}
 ||\hat{\bz}_{t,i}^\where - \bz_{t,j}^\where||_1 
                    &+ ||\hat{\bz}_{t,i}^\depth - \bz_{t,j}^\depth||_1 \nn\\
                    &- ({{\hat{\bz}_{t,i}}^{\pres}})^{\bz_{t,j}^{\pres}} {(1 - {\hat{\bz}_{t,i}}^{\pres})}^{1 - \bz_{t,j}^{\pres}}.\nn 
\end{align}
To keep all the terms commensurate, we scale $\textbf{z}^\where$ to be between 0 and 1 and use a standard normal distribution for the prior of $\textbf{z}^\depth$.
In scenes that are strictly 2D with no occlusion, such as our bouncing ball experiments, we do not include the depth term in this loss.

After applying the Hungarian algorithm with this cost matrix, we obtain a permutation matrix $P_t$.
We then left multiply $P_t$ with $\bz_t$ to obtain the aligned latents $\tilde{\bz}_t$.
This process is depicted in Figure \ref{fig:model}b.
Note that since we pre-train the encoder and freeze the weights when training the transformer, this alignment operation is not part of the computational graph. Thus, we are able to leverage the non-differentiable Hungarian algorithm to obtain the permutation matrix.

We should also mention that while we leverage the object positions with the Hungarian algorithm in our architecture, other alignment strategies, e.g. incorporating appearance information and learning the permutation matrix, may be used as well.
We leave this investigation for future work.


    

\subsubsection{Object-Wise Loss}
After aligning the objects over all timesteps, we are then able to train the transformer using the object-wise next-step prediction loss: $\mathcal{L_\text{object}}(\hat{\bz}_{t,k}, \tilde{\bz}_{t,k}) =$
\begin{align}
&||\hat{\bz}_{t,k}^\what - \tilde{\bz}_{t,k}^\what||_1 
+ \beta_\where ||\hat{\bz}_{t,k}^\where - \tilde{\bz}_{t,k}^\where||_1 \nn\\
&+ \beta_\depth ||\hat{\bz}_{t,k}^\depth - \tilde{\bz}_{t,k}^\depth||_1 \nn
- \beta_\pres[{\tilde{\bz}_{t,k}^\pres} \log (\hat{\bz}_{t,k}^\pres)\\ 
&+ ({1 - \tilde{\bz}_{t,k}^\pres}) \log (1 - \hat{\bz}_{t,k}^\pres)]\nn.
\end{align}
$\beta_\where$, $\beta_\depth$, and $\beta_\pres$ are hyperparameters used to control the contribution of each loss term.

\section{Experiments}

\textbf{}\textbf{Goal.} In our experiments, we seek to answer the following questions: (1) Can the model capture complex long-term spatiotemporal dependencies of the scene? (2) Can the model be effective at video generation? (3) Can the model provide good representations to use in downstream tasks? (4) How effective are our design choices (i.e., object-centric representations, transformer for dynamics prediction, scene prediction vs. autoregressive component prediction)?

\textbf{Datasets.} We evaluate \ours on a series of bouncing ball datasets designed to test jointly the long-term dependency, object interaction dynamics, and generation aspect of our model. We also evaluate on the CATER dataset \citep{cater}, a video-understanding benchmark that requires long-term temporal reasoning.

\textbf{Baselines.} For the bouncing balls dataset, we compare our model with GSWM, the previous RNN-based state-of-the-art for generation in this dataset. We further test against the following ablations of \oursns:

\begin{itemize}
    \item \textbf{\rnngnnns}: To test the effectiveness of using a transformer for temporal and interaction modeling, we replace the transformer in OCVT with an LSTM for temporal modeling and a GNN for interaction modeling. This model is conceptually similar to GSWM, except the encoder and decoder are pre-trained.
    \item To test the effectiveness of object-level tokenization, we replace the object-centric VAE with two other choices. (1) \textbf{Single-Vector Video Transformer} (\textbf{\vgptns}): The latent here is a single distributed vector representation for each image, tokenizing the video at the image level.
    (2) \textbf{\convgptns}: The latent here is a 4x4 grid of convolutional feature map cells. The grid is flattened and then used as input to the transformer. Similar to \oursns, we modify the transformer decoder causal mask to allow cells at the same timestep to attend to each other. The outputs of the transformer would be the updated states for each input cell at the next timestep.
    Note that we do not compare with a pixel-level tokenization scheme because of the large amount of compute required for such a model.
    \item \textbf{\vsortns}: To test our design decision of generating an entire image in one forward pass of the transformer and the importance of aligning objects between frames, we sort the objects based on the position from the top left of the image to the bottom right of the image and then predict each object autoregressively in the sorted order.
    \item \textbf{\convgptarns}: We use convolutional feature map cells as input to the transformer and predict each cell autoregressively in raster scan order. The last cell of an image predicts the first cell of the next image.
\end{itemize}

For CATER, we compare with Hopper \citep{hopper}, \citep{monetbert}, and OPNet \citep{opnet}.


\subsection{Bouncing Balls}

In this dataset, four colored balls bounce around in a frame.
Balls bounce off the walls and each other upon interaction.
Each ball is one of five colors and each color $k$ has an associated ordinal number $i_k$: $i_{blue} = 0$, $i_{red} = 1$, $i_{yellow} = 2$, $i_{violet} = 3$, $i_{cyan} = 4$.
If we denote the color of ball $o$ at frame $t$ by $c_{o, t}$, then when a ball hits the wall at frame $t$, it changes color according to the following formula: ($c_{o, t} + c_{o', t'}) \mod 5$.
Here $t'$ is the timestep of a previous interaction and $o'$ is the ball $o$ interacted with at time $t'$.
We test 4 different settings for $t'$ in our experiments.
The Mod1 dataset sets $t'$ to be the frame of the most recent interaction for object $o$.
The Mod2 and Mod3 datasets set $t'$ to be the frame of the second and third most recent interaction of object $o$, respectively.
The Mod1234 dataset uses a different $t'$ depending on the wall object $o$ is interacting with at time $t$.
If the ball interacts with the left wall, $t'$ is set to be the frame of the most recent interaction.
If the ball interacts with the top wall, $t'$ is set to be the frame from the second most recent interaction, and so on.

Each of these datasets is progressively more difficult in that they require longer-term and more complex (in the case of Mod1234) dependencies to be able to accurately predict future frames.
In order for a model to do well on these tasks, it needs to learn both the physical dynamics of the bouncing balls as well as the pattern of the color changes, which requires modeling of long-term dependencies, since the interactions may happen many timesteps in the past.
We evaluate the models under three settings: next-step prediction, long-term generation, and forced generation.

\begin{table}[tbp]
\caption{Average next-step prediction color change accuracy}
\label{table:next_step}
\begin{center}
\begin{small}
\begin{sc}
\begin{tabular}{lcccr}
\toprule
& Mod1 & Mod2 & Mod3 & Mod1234 \\
\midrule
GSWM         &   71.69   &   17.51   &  14.63    &  11.72    \\
\rnngnn      &   73.64   &   69.08   &  22.30    &  51.38    \\
\vgpt        &   37.53   &   18.23   &  11.96    &  29.47    \\
\convgpt     &   88.31   &   82.83   &  46.49    &  67.29    \\
\convgptar   &   8.70   &   4.20   &  3.25    &  6.10    \\
\vsort       &   78.70   &   76.99   &  54.49    &  64.97    \\
\midrule
\ours (ours) & \textbf{89.61} & \textbf{88.18} & \textbf{82.70} & \textbf{78.43} \\
\bottomrule
\end{tabular}
\end{sc}
\end{small}
\end{center}
\vskip -0.1in
\end{table}

\textbf{Next-Step Prediction.} In the next-step prediction setting, we task the models with predicting the next frame of the video given the history of ground truth frames.
To measure the accuracy of whether or not the models can correctly predict the color changes of the balls, we train a classifier that takes as input a patch from the ground truth image around each ball and predicts the color of that ball.
During test time, we use the reconstructed image and the ground truth positions to obtain a patch from the reconstructed image and use this classifier to determine the predicted colors of the balls for the different models.
Since the balls usually maintain their previous colors except for certain frames where they interact with the walls, this metric only contains the states where the color of the ball actually changes in the ground truth video.
Note that since we classify the patch of the reconstructed image where the ground truth ball is, performing well on this metric requires good performance on both ball dynamics prediction and ball color change prediction.

\begin{figure*}[t]
\centering
\includegraphics[width=\linewidth]{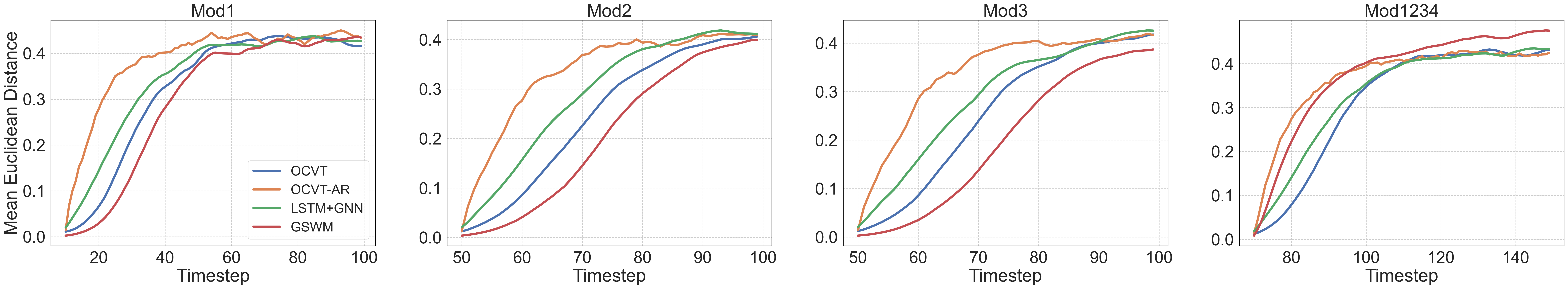}
\vspace{-7mm}
\caption{Generation mean Euclidean distance.}
\label{fig:med}
\end{figure*}

\begin{figure*}[t]
\centering
\includegraphics[width=\linewidth]{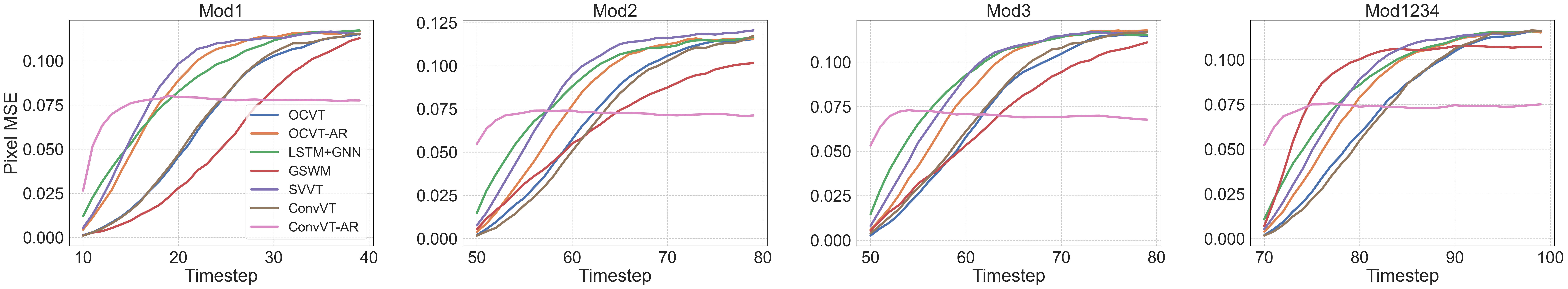}
\vspace{-7mm}
\caption{Generation pixel mean squared error}
\label{fig:pixel_mse}
\end{figure*}

The results are shown in Table \ref{table:next_step}.
We see that \ours achieves the highest accuracy across all four datasets.
\convgpt also performs well on the Mod1 dataset, but the accuracy degrades on the other datasets that require longer-term dependencies.
This suggests the effectiveness of using object-centric representations in modeling object-level spatiotemporal interactions.
Also, we notice a similar performance degradation for the RNN-based models (GSWM and \rnngnnns), indicating the limitations of using an RNN for modeling very long-term dependencies.
The models that autoregressively generate the scene (\vsort and \convgptarns) do not perform as well as their non-autoregressive counterparts.
\convgptar, in particular, seems to not be able to correctly model this task at all, achieving lower accuracy than random (0.2 with five colors).
This may be because the generation of a single image in these models requires multiple passes through the transformer and any prediction errors may be compounded.
Furthermore, \vsort requires the model to correctly learn the ordering of the objects based on their positions, which may not be an easy task.

\textbf{Long-Term Generation.} In the long-term generation setting, we provide a certain number of ground truth frames and ask the models to autoregressively predict a number of frames into the future.
For Mod1, we train on 20 frames, provide 10 ground truth frames, and ask the models to predict the next 90 frames.
For Mod2 and Mod3, we train on 50 frames, provide 50 ground truth frames and ask the models to predict the next 50 frames.
For Mod1234, we train on 70 frames, provide 70 ground truth frames and ask the models to predict the next 80 frames.


Measuring performance of long-term generation in this setting is difficult since early errors compound and can lead to prediction errors later in the trajectory.
In order to obtain the full picture of generation quality, we need to measure both the physical trajectories of the balls as well as the accuracy in the color change predictions of the balls.
For the models that provide explicit object position information (\oursns, \vsortns, \rnngnnns, GSWM), we can evaluate the trajectory by calculating the mean Euclidean distance between the predicted ball positions and the ground truth positions over time.
For models that do not provide object position information (\vgptns, \convgptns, \convgptarns), we can measure the pixel mean-squared error between the reconstructed image and the original image.
However, the pixel MSE metric can be misleading because low values may not necessarily correlate with good predictions.
For example, a model that predicts balls in the wrong positions would lead to a higher (worse) pixel MSE than a model that predicts a blank image, even though the model that predicts the blank image would be objectively incorrect.
Nonetheless, this metric can still provide information about generation quality, especially in the early parts of the trajectory when the predictions are still close to the ground truth.

The results are shown in Figures \ref{fig:med} and \ref{fig:pixel_mse}.
To highlight the early part of the trajectory, where pixel MSE is the most informative, we plot the curves until they flatten out instead of to the end of the trajectory.
We notice that GSWM achieves the best mean Euclidean distance and pixel MSE in this setting.
This is not surprising because GSWM's object-RNN encodes only the trajectory of a ball, isolated from the dynamics of the other objects.
Since the prediction of the dynamics only depends on the last few timesteps, no modeling of long-term dependencies is needed.
Interaction is dealt with by a separate graph neural network and is also well separated from the trajectory modeling.
\oursns, on the other hand, needs to learn to perform the more complex operation of factoring out its own trajectory while considering interaction as well.
Moreover, because GSWM is trained end-to-end, it can capture dynamics information in its latent variables, which would help in making trajectory predictions.
Interestingly, we find that \ours can still learn the dynamics reasonably despite this, outperforming the remaining baselines.
Furthermore, we note that GSWM's training procedure requiring a curriculum makes it take four times longer to converge than \oursns.
This can also lead to unstable training, especially for longer trajectories, as evidenced by the Mod1234 experiment where GSWM does not learn the dynamics well.

\begin{figure*}[h!]
\centering
\includegraphics[width=0.9\linewidth]{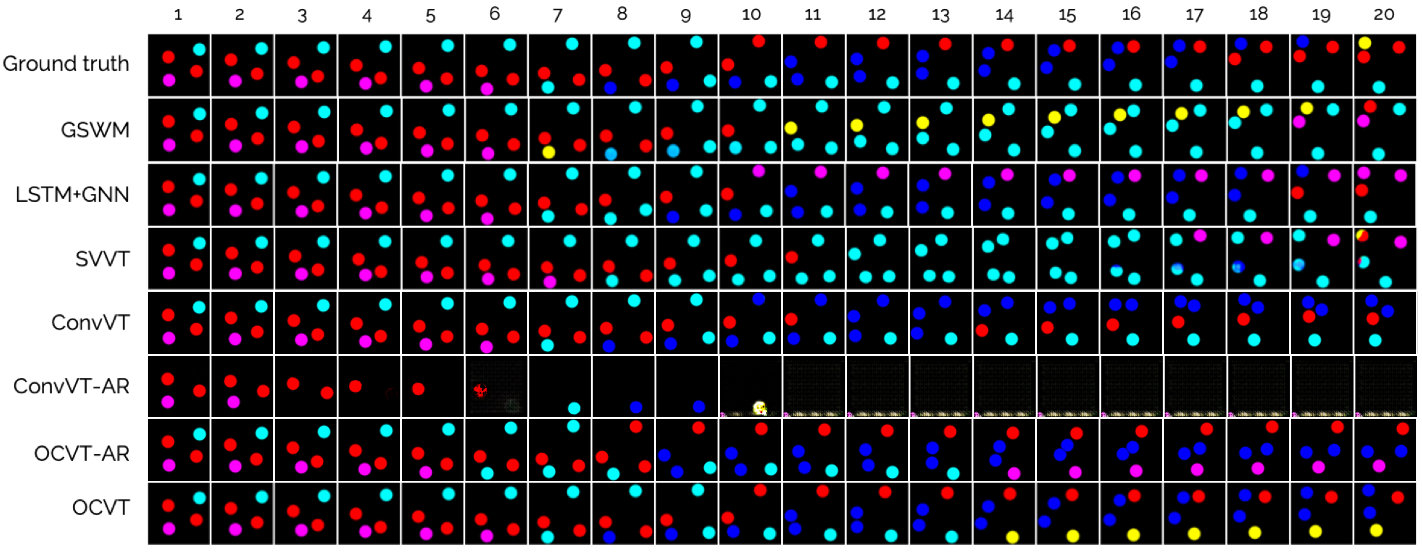}
\caption{Qualitative generation results for Mod1 dataset. Given the first 10 frames of the video, each model predicts the next 90 frames. The first 20 predicted frames are shown.}
\label{fig:gen_qual}
\end{figure*}

\convgpt performs similarly to \ours in terms of pixel MSE, although as we see in the qualitative results below, it makes mistakes predicting the color change. We also see that \convgptar flattens out quickly to a lower value than the other curves.
This is actually a result of the poor generation when the model incorrectly remove balls from the scene (see the qualitative results below).
However, since the pixel MSE between the ground truth and a blank image is around 0.06, this curve saturates at a lower value than the other curves. Lastly, we notice that \vsort performs worse than \ours across all the datasets, implying that autoregressive generation is not beneficial for of long-term generation.
As mentioned previously, this is likely due to the compounding errors that result from autoregressive object generation per image. 

\textbf{Qualitative Analysis.} Figure \ref{fig:gen_qual} shows the long-term generation results for the different models on the Mod1 dataset (see Supplementary for additional results).
The first 20 predicted steps are shown in the figure.
We see that GSWM predicts the locations of the balls fairly close to the ground truth, but makes several errors when predicting the color change of the balls.
For example, in frame 7, it incorrectly predicts that the violet ball should change color to yellow instead of cyan.
Similarly, \rnngnn and \convgpt incorrectly predict the color of the ball (eg. the violet ball at frame 10 for \rnngnn and the blue ball at frame 10 for \convgpt) while also having worse dynamics prediction than GSWM.
\vgpt has even worse dynamics than the other models and also makes predictions of balls with mixed colors.
\convgptar predicts missing balls after several frames resulting in a plateau at a lower pixel MSE than other models, even though the generation is clearly incorrect.
\vsort predicts color changes earlier than the ground truth frames because of inaccurate dynamics as well as incorrect color changes later in the trajectory (eg. violet ball in frame 14).

\oursns's dynamics prediction is relatively accurate compared to the other models (except for GSWM), but the slight difference in trajectory also causes the model to predict the color change of the yellow ball in frame 14 earlier than in the ground truth video.
Note, however, that the color change here is correct.
The ball is previously cyan, which has an ordinal value of 4, and following this ball earlier in the trajectory, we see the last interaction was with the violet ball in frame 4.
The ordinal value of violet is 3, so the expected color of the ball when it interacts with the wall at frame 14 is yellow, which has an ordinal value of 2.
Therefore, even though the color change occurred at an incorrect frame, the color change prediction is actually correct given the predicted dynamics.
This analysis also illustrates the difficulty of measuring generation quality in this setting where errors early in the trajectory compound.



\textbf{Forced Generation.}
Since the trajectories of the balls in the long-term generation setting may deviate from the ground truth (and the deviation compounds over time), we cannot easily evaluate whether or not the color change of the balls is correct.
To address this, we introduce a forced generation setting where we can perform this measurement.
For the models that provide explicit location and size information via the $\bz_\where$ latent, we can force the balls to follow the ground truth trajectories by directly manipulating $\bz_\where$ during prediction.
That is, during the prediction of the next frame, we manually set the $\bz_\where$ to be equal to the ground truth location and size of each ball.
All other latents are generated from the previous frame. 
This allows us to enforce the ground truth trajectory of the balls and measure how well the models learn the color change dynamics during long-term generation.

\begin{figure*}[t]
\centering
\includegraphics[width=\linewidth]{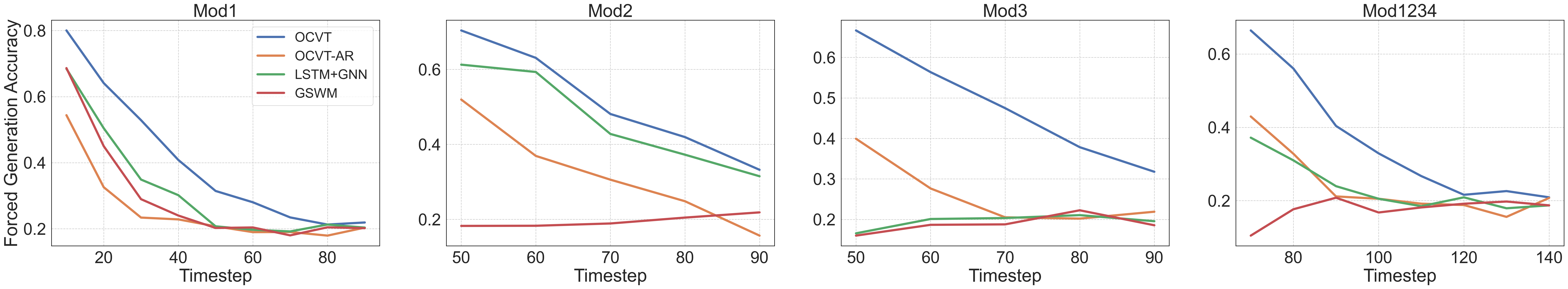}
\vspace{-7mm}
\caption{Forced Generation Accuracy}
\label{fig:accuracy}
\end{figure*}

We calculate the accuracy using the same classifier as in the next-step prediction setting and plot the resulting color change accuracy over time in Figure \ref{fig:accuracy}.
We see that \ours outperforms all the other baselines in this setting, indicating that our model is better able to learn the color change dynamics of the videos.
Even though GSWM is able to predict future object positions well, we notice that it does not perform well in this forced generation setting, with an accuracy of around 0.2 for Mod2, Mod3, and Mod1234, which is around the same accuracy of random guessing.
Similarly, \rnngnn also performs poorly in this setting, suggesting that the use of the transformer is beneficial for modeling these long-term dependencies.
While \vsort achieves higher accuracy than the RNN-based baselines, it does not surpass \oursns.
This may again be due to the compounding errors from the autoregressive generation.

\subsection{CATER Snitch Localization}
CATER \citep{cater} is a 3D dataset that consists of videos of objects moving around in a scene.
The objects lie on a 6x6 grid with the origin in the center of the scene.
The initial number and placement of objects is chosen randomly and objects also move and potentially cover other objects randomly as well.
Objects that cover other objects move together until they are uncovered.
The flagship task for this dataset is a snitch localization task, that requires predicting the location of the golden snitch (one of the objects that are present in all scenes) at the end of the video.
The snitch may be covered by another object (which may also be covered by other objects, and so on) so it may not be visible at the last frame and may have last been visible many frames prior to the last frame.
Therefore, this task requires a model to understand the effect of different actions in the environment and reason about them.
The final location is quantized into the 6x6 grid and the problem is set up as a single label classification task.

In order to handle this task, we modify our model to include a CLS token with a learned embedding as input to the transformer. An attention mask is added so that the latents $\textbf{z}$ cannot attend to the CLS token, but the CLS token can attend to all the latents at all timesteps. The output of the transformer corresponding to the CLS token is then used in an MLP to predict the final snitch location and a cross-entropy loss is used with the ground truth snitch locations. We pre-train the transformer and then fine-tune the entire model with the snitch localization objective. 

Table \ref{table:cater} shows the results of our experiments. The Top 1 and Top 5 accuracy, as well as the final L1 distance between the prediction and the ground truth location, are reported.
We compare \ours with several previous attempts to solving this problem, including two approaches that also combine object-centric representations with transformers, Hopper \citep{hopper} and \citep{monetbert}.
Hopper requires several auxiliary losses to perform well on this task including a loss that requires knowing the first and second movements of the snitch.
\citet{monetbert} use a BERT-like bidirectional masking scheme and achieves the best performance by adding an L1 loss to the objective for the location of the snitch.
Our model outperforms these baselines in Top 1 and Top 5 accuracy without the use of any auxiliary losses.
With the addition of the L1 auxiliary loss, our model also achieves the lowest L1 distance.
This demonstrates that our model can learn good representations of the scene to be used in downstream tasks.

\vspace{-3mm}
\begin{table}[tbp]
\caption{CATER results}
\label{table:cater}
\vskip 0.15in
\vspace{-3mm}
\begin{center}
\begin{small}
\begin{sc}
\begin{tabular}{lcccr}
\toprule
& Top 1 $\uparrow$ & Top 5 $\uparrow$ & L1 $\downarrow$\\
\midrule
Ding et al           &  70.6           &  93.0            & 0.53   \\
Ding et al w/ L1     &  74.0           &  94.0            & 0.44   \\
Hopper               &  73.2           &  93.8            & 0.85   \\
OPNet                &  74.8           &  -               & 0.54   \\
\midrule
\ours (ours)         &  \textbf{76.0}           &  94.4            & {0.45} \\
\ours w/ L1 (ours)   &  75.9           &  \textbf{95.3}            & \textbf{0.39} \\
\bottomrule
\end{tabular}
\end{sc}
\end{small}
\end{center}
\vskip -0.3in
\end{table}







\section{Conclusion}
We proposed \oursns, a generative video transformer that leverages the recent advances in unsupervised object-centric representation learning.
Our model is able to generate future frames of videos with complex long-term dependencies and learn representations that are useful in downstream tasks.
This study shows that given an appropriate representation, using \textit{objects as visual words} can be a reasonable inductive bias for tokenizing a video.
While our model uses SPACE to obtain object representations and leverages the explicit position information to align objects, future work may involve improving the object-centric representations to work for more complex, real-world scenes.
Furthermore, it would be interesting to investigate alignment strategies with representations without explicit position latents.


\section*{Acknowledgements}
The authors would like to thank the anonymous reviewers for constructive comments.

\bibliography{refs}
\bibliographystyle{icml2021}

\input{supplementary}

\end{document}

%% file: supplementary.tex











\twocolumn[
\icmltitle{Supplementary Material for \\``Generative Video Transformer: Can Objects be the Words?''}
{\begin{figure}[H]
\setlength{\linewidth}{\textwidth}
\setlength{\hsize}{\textwidth}
\centering
\includegraphics[width=0.9\linewidth]{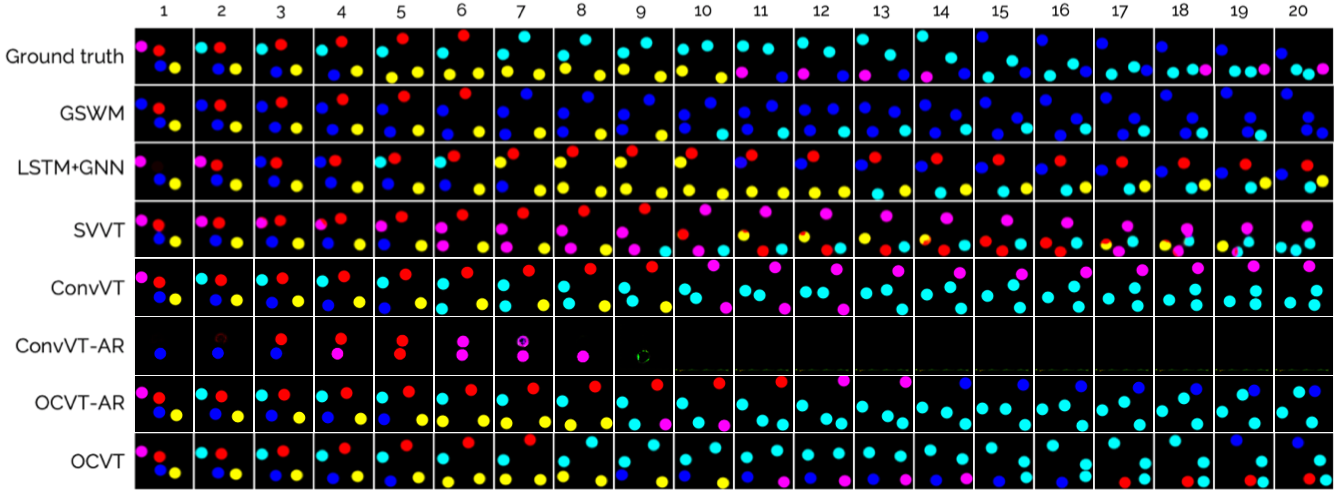}
\caption{Qualitative generation results for Mod2 dataset. Given the first 50 frames of the video, each model predicts the next 50 frames. The first 20 predicted frames are shown.}
\label{fig:gen_qual_mod2}
\end{figure}}
]

\begin{figure*}[htp]
\centering
\includegraphics[width=0.9\linewidth]{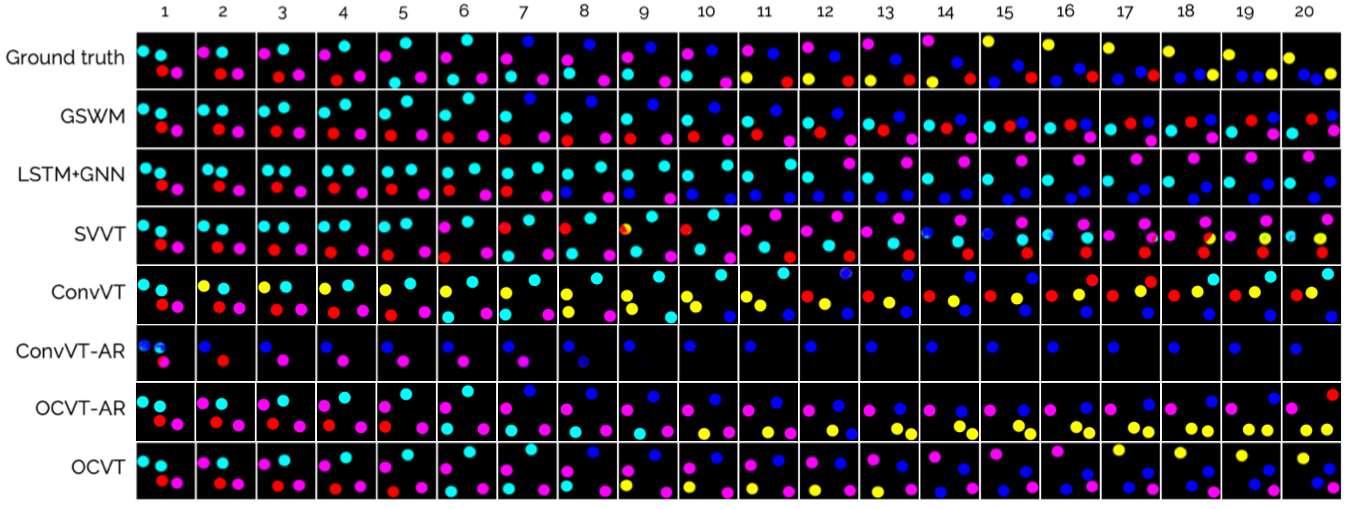}
\caption{Qualitative generation results for Mod3 dataset. Given the first 50 frames of the video, each model predicts the next 50 frames. The first 20 predicted frames are shown.}
\label{fig:gen_qual_mod3}
\end{figure*}

\begin{figure*}[htp]
\centering
\includegraphics[width=0.9\linewidth]{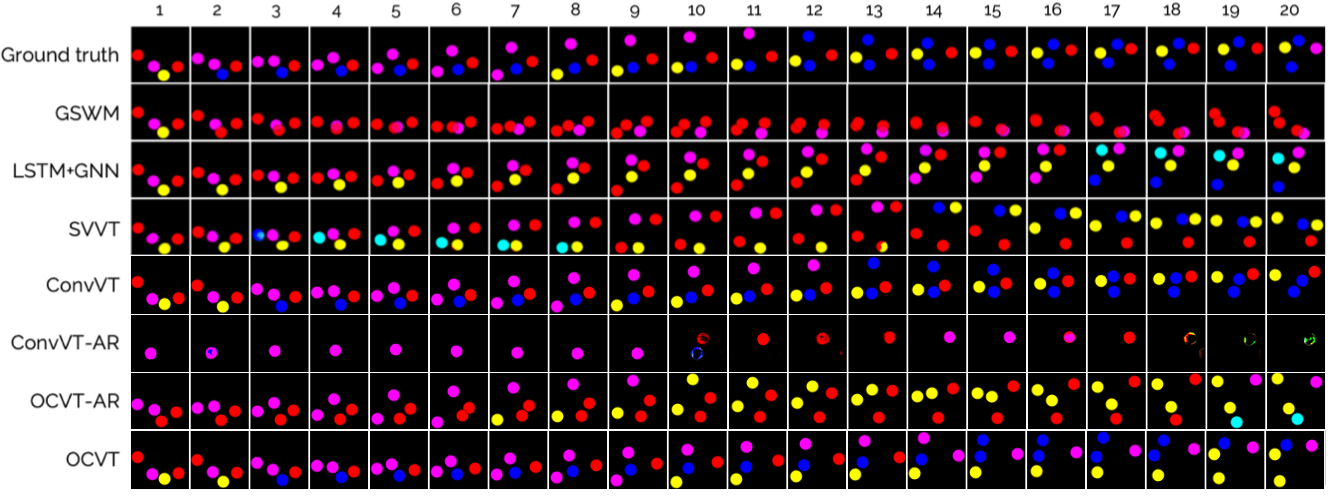}
\caption{Qualitative generation results for Mod1234 dataset. Given the first 70 frames of the video, each model predicts the next 80 frames. The first 20 predicted frames are shown.}
\label{fig:gen_qual_mod1234}
\end{figure*}

\appendix

\section{Additional Results}
\subsection{Qualitative Generation Results}

Figures \ref{fig:gen_qual_mod2}, \ref{fig:gen_qual_mod3}, \ref{fig:gen_qual_mod1234} show qualitative generation results for the Mod2, Mod3, and Mod1234 datasets.

\subsection{Attention Analysis}

 \begin{figure}[h]
 \centering
 \includegraphics[width=0.9\linewidth]{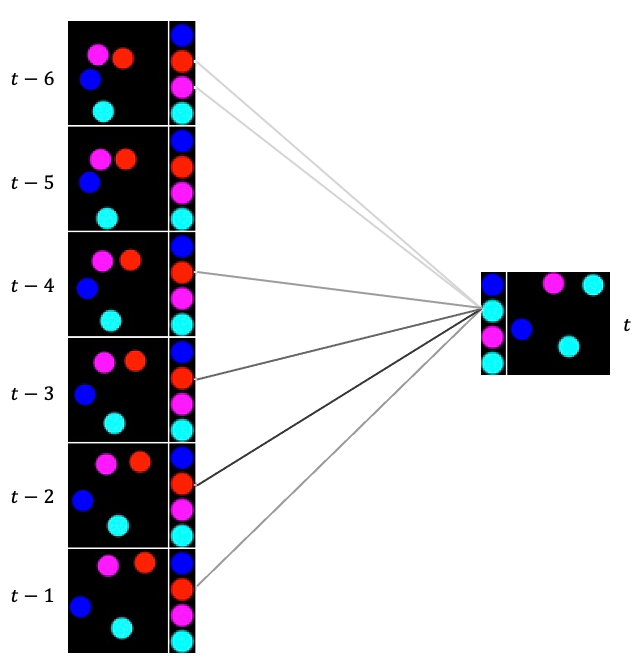}
 \caption{Attention strength at $7^\text{th}$ layer during color change.}
 \label{fig:attn_change}
 \end{figure}
 
Figure \ref{fig:attn_change} shows an example of the attention weights in the transformer when predicting the last timestep of a sequence in the Mod1 dataset.
The right hand side shows the balls at timestep $t$ and the left hand side shows the balls for the 6 timesteps prior.
The darker the shade of gray, the stronger the weight.
At this particular timestep, the color of the top right ball changes from red to cyan.
We see that the strongest attention weights are to the same ball in the previous frames as well as the violet ball several frames prior, which is the ball that last interacted with this ball.
This makes intuitive sense because the positions of the same ball in the last few frames are important in predicting to updated location of the ball at the next timestep.
In order to correctly predict the color change of the ball, it must also attend to the ball that it most recently interacted with.

\subsection{End-to-End Training}
We also evaluated \ours in a setting where we train the entire model end-to-end instead of freezing the parameters of the encoder and decoder while training the transformer.
This is done under two settings: (a) training the model completely from scratch end-to-end and (b) using a pre-trained encoder and fine-tuning the model end-to-end.
We achieve a next-step change accuracy of 82.21\% for (a) and 76.95\% for (b) for the Mod1 dataset.
While this end-to-end training does not outperform our best pre-trained model, end-to-end training may be beneficial in certain scenarios since the encoder can incorporate temporal information from the scene.
We leave this investigation for future work.

\section{Implementation Details}

\subsection{Model Architecture}
For the foreground image encoder, we use a ResNet18 \citep{resnet}.
For $(H,W)=(4,4)$, we apply an extra pair of $3 \times 3$ convolutions with stride 1 to get the appropriate dimensions per grid cell (see Hyperparameters in the next section).
For $(H,W)=(8,8)$, we remove the last ResNet block and then apply the pair of convolutions.
To obtain $\bz_t$, each cell is run through a 3-layer fully convolutional network with ReLU activation and group normalization \citep{groupnorm}.
After the final layer, we apply softplus to compute standard deviations of the Gaussian distributions for $\bz_t^\where$, $\bz_t^\what$, $\bz_t^\depth$.
For $\bz_t^\pres$, we apply the sigmoid function and use the Gumbel-Softmax \citep{gumbel} relaxation to model a Bernoulli random variable.
For the foreground image decoder, we use a 6-layer sub-pixel convolutional network \citep{pixelshuffle} with group normalization in the intermediate layers.

For the background image encoder, we use a 4-layer convolutional network with CELU activation \citep{celu} and group normalization followed by a final linear layer.
For the background image decoder, we use a 6-layer convolutional network, each consisting of 2D bilinear upsampling followed by a convolution with leaky ReLU activation.

For the transformer, we use a linear layer to obtain the desired dimensions for the transformer input (see Hyperparameters).
The output of the transformer runs through a single hidden layer MLP with ReLU activation to obtain the next step predictions.

\subsection{Hyperparameters}
We provide the hyperparameters used in our experiments in Tables \ref{tab:hyp_balls} and \ref{tab:hyp_cater}.

\begin{table}[h!]
    \caption{List of Hyperparameters for Bouncing Ball Datasets}
    \centering
    \begin{tabular}{llll}
    \toprule
    Description                    & Value \\
    \midrule         
    Image Size                     & (64,64) \\
    Grid size $(H, W)$             & (4,4) \\
    Dimension per grid cell        & 128 \\
    Dimension of $\bz^\what$       & 16 \\
    Dimension of $\bz^{bg}$        & None \\
    Foreground Variance            & 0.2 \\
    Background Variance            & None \\
    Gumbel-Softmax Temp. for $\bz^\pres_t$ & 0.01 \\
    $\beta_\where$ & 20 \\
    $\beta_\depth$ & 0 \\
    $\beta_\pres$ & 1 \\
    $\beta_\what$ & 4 \\
    Dimension of transformer input & 360 \\
    Feedforward dimension in transformer & 256 \\
    Number of heads & 8 \\
    Number of transformer layers & 15 \\
    \bottomrule
    \end{tabular}    
\label{tab:hyp_balls}
\end{table}

\begin{table}[h!]
    \caption{List of Hyperparameters for CATER Datasets}
    \centering
    \begin{tabular}{llll}
    \toprule
    Description                    & Value \\
    \midrule         
    Image Size                     & (64,64) \\
    Grid size $(H, W)$             & (8,8) \\
    Dimension per grid cell        & 128 \\
    Dimension of $\bz^\what$       & 64 \\
    Dimension of $\bz^{bg}$        & 64 \\
    Foreground Variance            & 0.05 \\
    Background Variance            & 0.2 \\
    Gumbel-Softmax Temp. for $\bz^\pres_t$ & 0.01 \\
    $\beta_\where$ & 50 \\
    $\beta_\depth$ & 1 \\
    $\beta_\pres$ & 1 \\
    $\beta_\what$ & 1 \\
    Dimension of transformer input & 360 \\
    Feedforward dimension in transformer & 256 \\
    Number of heads & 6 \\
    Number of transformer layers & 15 \\
    \bottomrule
    \end{tabular}    
\label{tab:hyp_cater}
\end{table}

\section{Dataset and Experiment Details}
\subsection{Bouncing Balls}
In all the bouncing ball datasets, we have 20,000 videos for training, 200 videos for validation, and 200 videos for testing.
For the Mod1, Mod2, and Mod3 datasets, each video has an episode length of 100 frames.
For the Mod1234 dataset, each video has an episode length of 150 frames.
This longer episode length is to allow for a sufficient number of interactions (up to 4 for this dataset) in the videos.
We choose the best model based on the change accuracy on the validation set and then use this model on the test set for evaluation.
All models are trained to convergence measured by the plateauing of the change accuracy on the validation set.

\subsection{CATER}

This dataset consists of 3,080 videos for the training set, 770 videos for the validation set, and 1650 videos for the test set.
Each video frame is reshaped to 64x64 pixels.
Each video originally has 300 frames and we randomly sample 50 frames for training.
For validation and testing, we take every sixth frame for a total of 50 frames.
We choose the best model based on the best Top 5 Accuracy for the snitch localization task on the validation set and then use this model on the test set for evaluation.


